\pdfoutput=1
\documentclass[11pt]{article}

\usepackage{acl}

\usepackage{times}
\usepackage{latexsym}

\usepackage[T1]{fontenc}
\usepackage[utf8]{inputenc}

\usepackage{microtype}

\usepackage{inconsolata}

\usepackage{graphicx}

\usepackage{amsmath}
\usepackage{booktabs}
\usepackage{amssymb}% http://ctan.org/pkg/amssymb
\usepackage{pifont}% http://ctan.org/pkg/pifont

\usepackage{multirow}
\usepackage{multicol}
\usepackage{adjustbox}
\usepackage{arydshln}
\usepackage{tcolorbox}
\usepackage{hhline} % For \dhashline
\usepackage{enumitem} % For better list customization

\definecolor{myblue}{HTML}{4E84C4}
\definecolor{myred}{HTML}{B02418}
\definecolor{mygreen}{HTML}{34692E}
\definecolor{myorange}{HTML}{DA7842}
\definecolor{paperblue}{HTML}{077dea}
\definecolor{babyblue}{HTML}{E3EDF7}

\usepackage{amssymb}

\newcommand{\coloredalpha}{\textcolor{paperblue}{\alpha}}
\newcommand{\coloredgamma}{\textcolor{paperblue}{\gamma}}
\newcommand{\coloredsigma}{\textcolor{paperblue}{\sigma}}
\newcommand{\coloreddelta}{\textcolor{paperblue}{\delta}}

\newcommand{\coloredtau}{\textcolor{paperblue}{\tau}}

\title{Self-Reasoning Language Models: Unfold Hidden Reasoning \\ Chains with Few Reasoning Catalyst}

\author{Hongru Wang$^{\coloredalpha}$, Deng Cai$^{\coloredgamma}$, Wanjun Zhong$^{\coloredgamma}$, Shijue Huang$^{\coloredgamma}$, \\ \bf Jeff Z. Pan$^{\coloreddelta}$, Zeming Liu$^{\coloredsigma,}$\textsuperscript{\rm{\ding{41}}}, Kam-Fai Wong$^{\coloredalpha, \coloredtau,}$\textsuperscript{\rm{\ding{41}}} \\
  $^{\coloredalpha}$The Chinese University of Hong Kong,
  $^{\coloredgamma}$ByteDance \\
  $^{\coloreddelta}$The University of Edinburgh,
  $^{\coloredsigma}$Beihang University \\
  $^{\coloredtau}$MoE Key Laboratory of High Confidence Software Technologies \\
  \texttt{zmliu@buaa.edu.cn}, \texttt{\{hrwang, kfwong\}@se.cuhk.edu.hk} }

\begin{document}
\maketitle
\begin{abstract}

Inference-time scaling has attracted much attention which significantly enhance the performance of Large Language Models (LLMs) in complex reasoning tasks by increasing the length of Chain-of-Thought. These longer intermediate reasoning rationales embody various meta-reasoning skills in human cognition, such as reflection and decomposition, being difficult to create and acquire. In this work, we introduce \textit{Self-Reasoning Language Model} (SRLM), where the model itself can synthesize longer CoT data and iteratively improve performance through self-training. By incorporating a few demonstration examples (i.e., 1,000 samples) on how to unfold hidden reasoning chains from existing responses, which act as a reasoning catalyst, we demonstrate that SRLM not only enhances the model's initial performance but also ensures more stable and consistent improvements in subsequent iterations. Our proposed SRLM achieves an average absolute improvement of more than $+2.5$ points across five reasoning tasks: MMLU, GSM8K, ARC-C, HellaSwag, and BBH on two backbone models. Moreover, it brings more improvements with more times of sampling during inference, such as absolute $+7.89$ average improvement with $64$ sampling times, revealing the in-depth, diverse and creative reasoning paths in SRLM against the strong baseline.

\end{abstract}

\section{Introduction}

Recent studies have demonstrated that inference-time scaling \cite{openai2024_o1} effectively improves performance of Large Language Models (LLMs) in various reasoning tasks, such as mathematics and complex question answering, by extending the length of Chain-of-Thought (CoT) \cite{wei2022cot} reasoning rationales \cite{qwq2024, deepseekai2025deepseekr1}. With longer CoTs during inference, these LLMs could explore more creative and diverse reasoning rationales while assembling various meta-reasoning skills observed in human cognition \cite{gigerenzer1991tools, johnson1988computer, dye2011reflection, tpe_cognitive}, such as reflection and decomposition \cite{tool_tut, wang2024selfdc, wang2025thoughtsplaceunderthinkingo1like}, resulting in better performance. However, the \textit{scarcity} of these longer CoT data remains a significant obstacle to advancing inference-time scaling, especially for instructions without verifiable answers in the general instruction-tuning data.

\begin{figure}
    \centering
     \includegraphics[trim={4.5cm 7.5cm 14cm 6cm}, clip, width=0.5\textwidth]{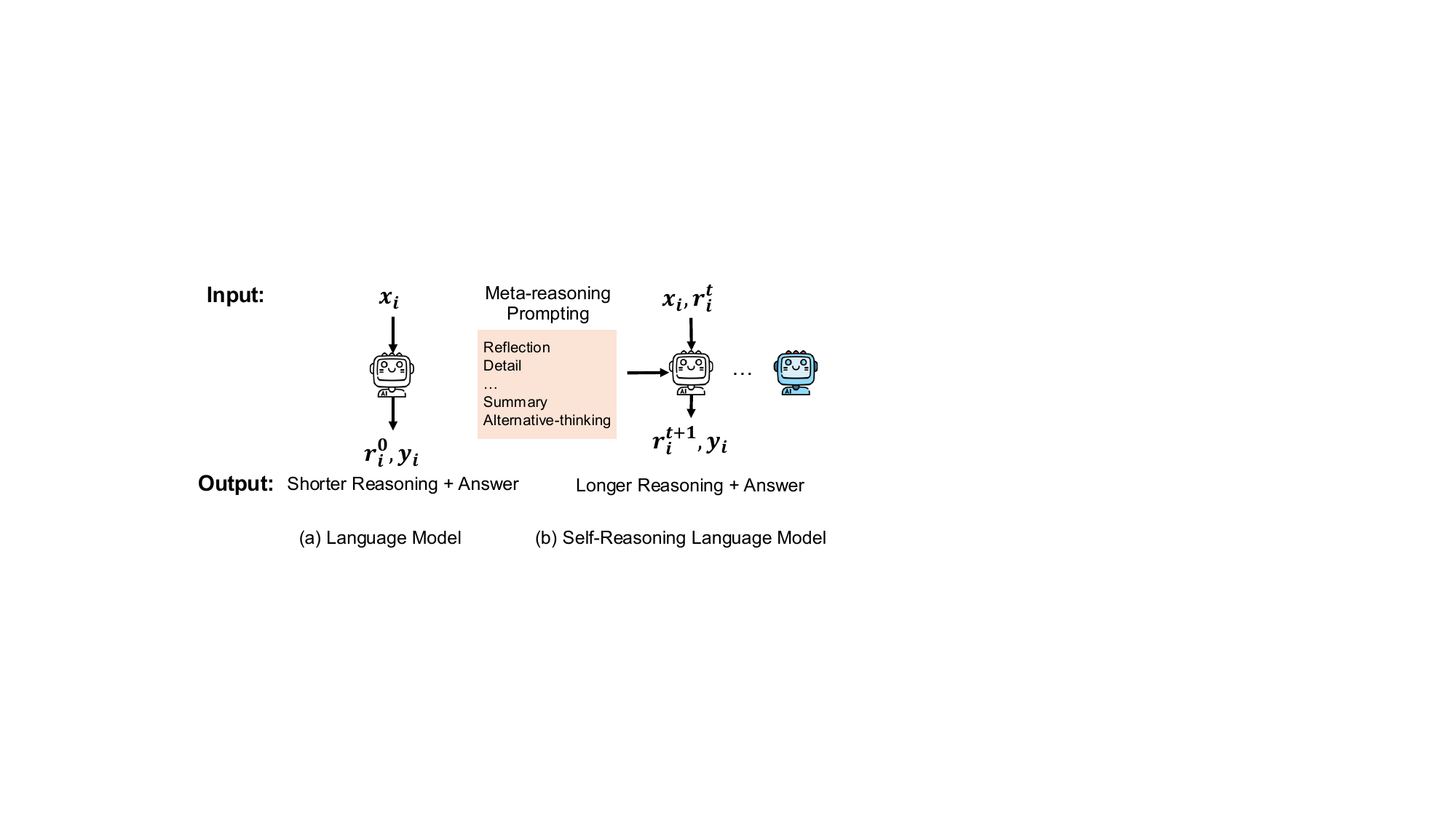}
    \caption{The comparison between (a) language model; and (b) self-reasoning language model where the later is capable to generate more longer reasoning chains alongside the final answer.}
    \label{fig:intro}
    \vspace{-4mm}
\end{figure}

Several prior studies have explored various approaches to get better CoT tuning datasets, where most of them utilize either the LLM itself \cite{wang2023self, huang-etal-2023-large, wang-etal-2023-cue} or external reasoning models \cite{li-etal-2024-selective, hu2024nileinternalconsistencyalignment} to generate (\textit{or refine}) new (\textit{or existing}) responses in the instruction-tuning dataset, leading to great improvements at various downstream tasks. Despite the effectiveness of these proposed methods, they still face several limitations. On the one hand, most of them focus on questions with verifiable answer such as math and code \cite{huang-etal-2023-large, deepseekai2025deepseekr1}, being infeasible for general instruction-tuning dataset. On the other hand, another line of work typically assumes access to more powerful models to refine each sample iteratively \cite{xu2023wizardlm, li-etal-2024-selective}. Such methods suffer from performance plateaus or even degradation \cite{ding2024mitigatingtailnarrowingllm} and are inherently constrained by the capability ceiling of the powerful model.

To this end, we present \textit{Self-Reasoning Language Models} (SRLM), which is capable to self-unfolding its own reasoning rationales and iteratively optimize itself, leading to enhanced overall capability. Specifically, we first create only few reasoning catalyst data that compose the demonstrations of how to enrich shorter CoT rationales into more longer and comprehensive CoT with the augmentation of various meta-reasoning skills. After incorporating the reasoning catalyst data with the original instruction-tuning data, the tuned model not only inherit the basic reasoning capabilities from the instruction-tuning dataset, but also learn how to refine reasoning simultaneously, resulting in \textit{Self-Reasoning Language Models}. Consequently, the SRLM can refine its own reasoning rationales at each iteration with the processing of reasoning expansion and selection. During this process, the model generates enriched reasoning rationale candidates for the same instructions in the original instruction-tuning dataset. These rationale pairs are then filtered and selected using three proposed selectors without any prior assumption about the instruction and answer. Finally, the newly selected instruction-tuning dataset is combined with the reasoning catalyst data to create the training data for the next iteration of SRLM, which is initialized from the same base model. To conclude, our contributions can be summarized as follows:

\begin{itemize}[leftmargin=*,topsep=1pt,itemsep=1pt]
    \item We introduce \textit{Self-Reasoning Language Models} (SRLM), which is capable to act as both: 1) a response generation model (i.e., \textit{learn to reason}); and 2) reasoning models to refine its own reasoning rationales (i.e., \textit{how to reason}).

    \item We present \textit{reasoning catalyst data}, function as demonstrations to guide the LLM to unfold hidden reasoning chains with the augmentation of various meta-reasoning skills such as reflection and decomposition.

    \item Our experiments demonstrate that incorporating a small portion of reasoning catalyst data (less than $0.02\%$) enables small-size SRLMs to generate higher-quality instruction-tuning datasets compared to those generated by GPT-4o, while also achieving better performance and more stable improvements across iterations on various reasoning benchmarks.
\end{itemize}

\section{Related Work}

\paragraph{Synthetic SFT Data.} Previous research on instruction tuning has primarily utilized human-annotated data to create large, high-quality datasets \citep{wang-etal-2022-super}, being both time-consuming and labor-intensive. Therefore, recent studies turn to leverage more advanced models to automatically generate instruction-tuning datasets \citep{wang2023self, peng2023instructiontuninggpt4, li-etal-2024-selective}. For instance, Magpie \citep{xu2024magpiealignmentdatasynthesis} leverage open-source models to generate more diverse user queries and responses by only feeding the left-side templates up to the position reserved for user messages as input. Moreover, there is another line of works focused on refining existing instruction-tuning data using powerful models, such as WizardLM \cite{xu2023wizardlm} and Reflection-tuning \cite{li-etal-2024-selective} and NILE \cite{hu2024nileinternalconsistencyalignment}. In detail, \citet{li-etal-2024-selective} utilize GPT-3.5 to reflect on existing instructions and responses while \citet{hu2024nileinternalconsistencyalignment} use GPT4 to capture the knowledge gap between internal parametric knowledge in LLMs and external world knowledge in instruction-tuning data.

\paragraph{Self-Improved LLMs.} Recent studies have demonstrated that LLMs can refine themselves using self-generated data, leading to enhanced performance and efficiency without human intervention, resulting in self-improving LLMs \cite{huang-etal-2023-large}. It is crucial to select high-quality samples from the vast amounts of generated data. There are different ways to acquire the quality signals of data samples: 1) LLMs \cite{yuan2024selfrewardinglanguagemodels}; 2) ranking or reward models \cite{dong2023raft, lu2024autopsv}; and 3) some automatic methods such as execution feedback \cite{haluptzok2022language} or self-consistency in multiple reasoning paths \cite{wang2023selfconsistencyimproveschainthought}. One distinct aspect of our work is that we are working on general instruction-following data optimization iteratively, where the answers are not verifiable like math and coding problems, making most of previous studies infeasible and unaffordable \cite{huang-etal-2023-large, li-etal-2024-selective}.

\begin{figure*}
    \centering
     \includegraphics[trim={3cm 5cm 3cm 4cm}, clip, width=1.0\textwidth]{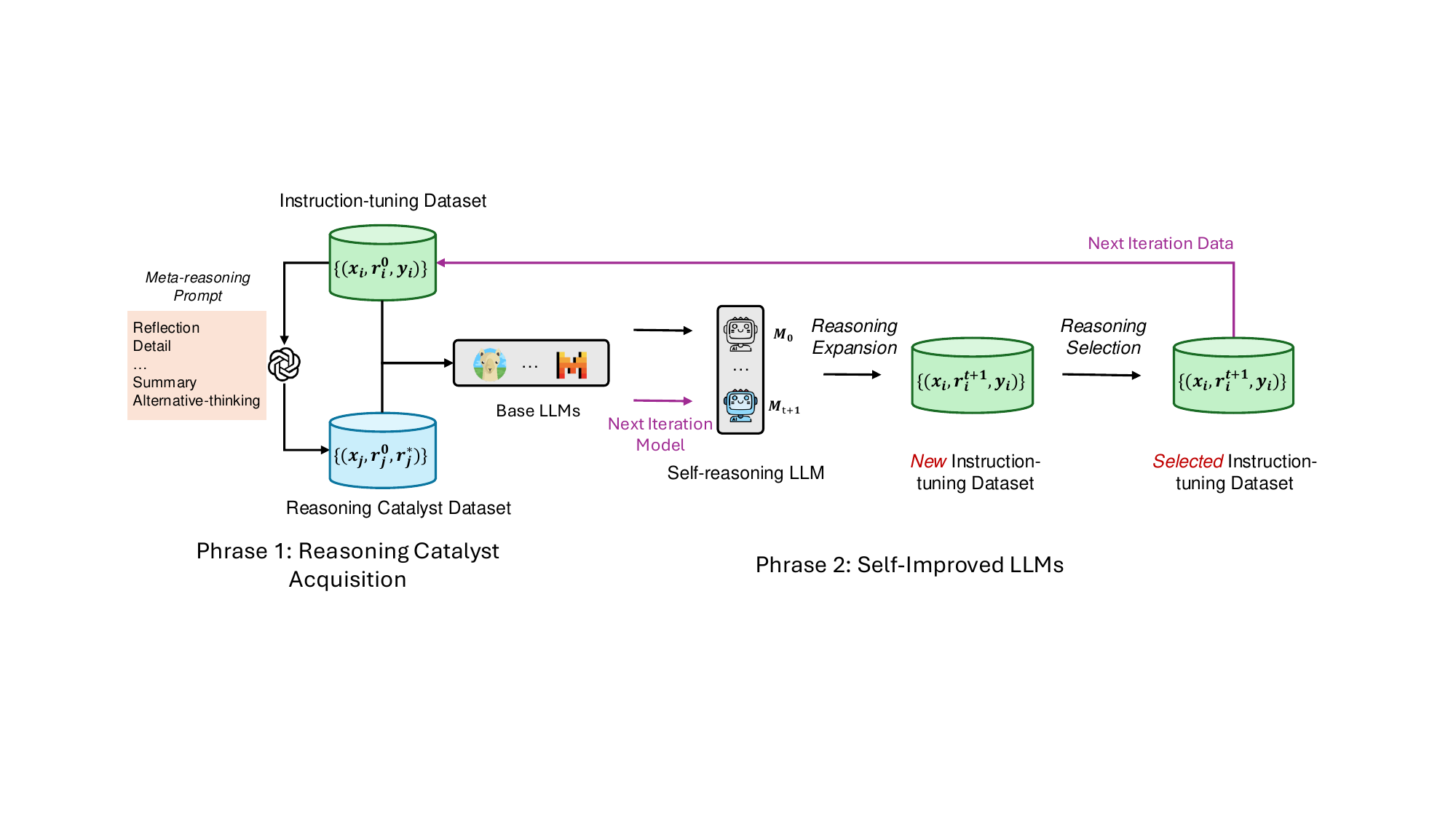}
    \caption{The framework of our proposed Self-Reasoning Language Models, which consists of two phrases.}
    \label{fig:ggl}
\end{figure*}

\section{Method}

\subsection{Task Formulation}

Given a large language model $\mathcal{M}$ and the original instruction-tuning  dataset $D^0 = \{(x_i, r^0_i, y_i)\}$ at time step $0$, where $x_i$ is the instruction, $r_i$ is the chain-of-thought rationale \cite{wei2022cot} \footnote{Note this is optional, take it as $y_i$ if it does not exist.} and $y_i$ represents the final answer in the $i_{th}$ sample. The goal of self-improved LLMs is to iteratively refine the instruction-tuning dataset from the time step $0$, leading to better dataset $D^t = \{(x_i, r^t_i, y_i)\}$ at time step $t$, ultimately resulting in enhanced performance at downstream tasks.

% The goal of the task is to iteratively enrich the reasoning processing $r^0_i$ from the start time, resulting in better instruction-tuning datasets in long CoT, leading to better performance.

\subsection{Framework}
As shown in Figure~\ref{fig:ggl}, there are two main phases in our method. In phase 1, we leverage existing LLMs to acquire seed data of reasoning augmentation, and function as reasoning catalyst to enlight the models to enrich existing reasoning rationales (\$\ref{reasoning_catalyst}). In phase 2, the model then is fine-tuned using both original instruction-tuning data and reasoning catalyst data, and iteratively refined using self-generated new rationales, resulting in self-improved models (\$\ref{self-improvedllms}).

% What we aim to do is expand the reasoning processing in $$a_0$$ iteratively to get $$a_1, a_2, ...$$, leading to more comprehensive and effective reasoning processing. Specifically, we divide our framework into two stages:

\subsubsection{Reasoning Catalyst Acquisition}
\label{reasoning_catalyst}

There is a Chinese proverb: ``\textit{It's better to teach a man to fish than to give him fish.}" Similarly, in the context of LLMs, rather than directly copying reasoning capabilities from existing instruction-tuning datasets, it is crucial to teach LLMs how to reason and refine their reasoning based on specific problems and existing reasoning rationales. To achieve this goal, we first carefully design meta-reasoning prompt \footnote{The prompt can be found in Table~\ref{table:meta_reasoning_prompt}.}, which is composed of different meta-reasoning skills such as \textit{reflection}, \textit{detail}, and \textit{alternative-thinking}, inspired by meta-reasoning theory in cognitive science \cite{zhou2024self}. Then we leverage these meta-reasoning skills to unfold hidden reasoning chains based on given instruction and previously reasoning rationales $r^0_{j}$, resulting in enriched reasoning rationales $r_{j}^*$ as follows:

\begin{equation}
    r_j^* = \mathcal{M}_{meta} (x_j, r_j^0)
\end{equation}

Here the $\mathcal{M}_{meta}$ could be human experts or any off-the-shelf models, $r_j^*$ starts with \textit{<thoughts>} and ends with \textit{</thoughts>}. All other meta-reasoning skills used in the thoughts also be clearly indicated in a similar manner for better understanding and analysis. Therefore, it can effectively fill in missing details within existing rationales while preserving their core semantic meaning. Moreover, it is desirable to introduce new rationales (i.e., using \textit{alternative-thinking}), correct errors in previous rationales (i.e., using \textit{reflection}), and other hidden reasoning chains with different skills. Thus, this process provides exceptional flexibility and robustness to address errors, misleading reasoning paths, and unclear solutions in the previous reasoning rationale. Since we aim to teach the model how to reason instead of directly telling them the correct reasoning paths for all instructions. Therefore, we only incorporate a few samples (only 0.02\% in the main experiment) as reasoning catalyst data to enforce the models' generalization to other unseen instructions.

\subsubsection{Self-Improved LLMs}
\label{self-improvedllms}

Given the original instruction-tuning dataset $D^t$ = $\{(x_i, r^t_i, y_i)\}$ ($t$ = 0 at the beginning) and fixed few reasoning catalyst data $R$ = $\{(x_j, r^t_j, r^*_j)\}$, the original base LLM $\mathcal{M}$\footnote{We note we always finetune the model start from base models instead of continual tuning.} is then tuned using both types of data to learn reasoning and how to reason simultaneously, resulting in new LLM $\mathcal{M}_t$. Then we enrich the reasoning rationales in original instruction-tuning datasets (i.e., \textit{iterative reasoning expansion}) and select better rationales to form the instruction-tuning dataset in the next step $t+1$ (i.e., \textit{iterative reasoning selection}), which completes the whole iteration.

\paragraph{Iterative Reasoning Expansion.} We directly prompt the $M_t$ model to generate enriched reasoning rationales given previous reasoning $r_j^t$ and $x_j$ for \textit{each sample} in the instruction-tuning dataset, distinguishing from reasoning catalyst acquisition.

\begin{equation}
\label{sample_eq}
    r_{i,m}^{t+1} = \mathcal{M}_{t} (x_i, r_i^t), \quad \forall m \in \{1, 2, \dots, N\}.
\end{equation}

where $N$ represents the number of sampling attempts, designed to enhance diversity and produce higher-quality output.

\begin{figure}
    \centering
     \includegraphics[trim={9cm 7cm 12cm 2cm}, clip, width=0.5\textwidth]{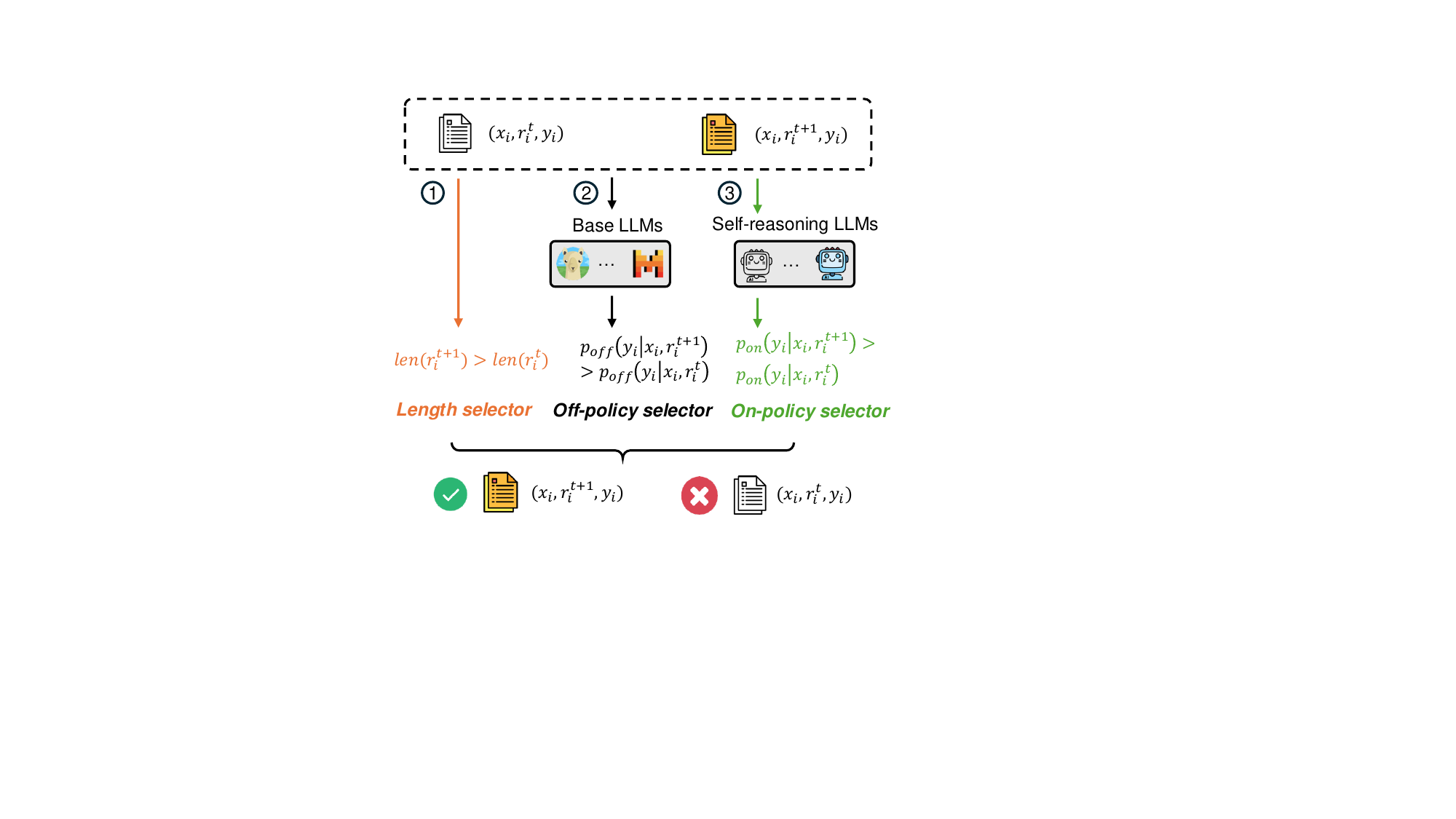}
    \caption{Three different types of reasoning selectors.}
    \label{fig:selector}
\end{figure}

\paragraph{Iterative Reasoning Selection.} It is important to determine which rationale is better for current instruction given the pair of $(x_i, r^t_i, y_i)$ and $(x_i, r^{t+1}_i, y_i)$. Thus we carefully design three different selectors to investigate corresponding effects as shown in Figure~\ref{fig:selector}:

\begin{itemize}[leftmargin=10pt]
    \item \textbf{Length Selector.} We consider a longer rationale to be better because it provides more detailed insights into the previous reasoning. Even if it contains errors or irrelevant information, it can be refined and corrected in subsequent iterations.

    \item \textbf{Off-policy Selector.} We leverage the \textit{original base model} to calculate the the probability of the answer $y_i$ given the instruction $x_i$ and different rationales, and then choose the rationale that has a higher chance of leading to the final answer. 

    \begin{equation}
        r^* = \arg\max_{j \in \{t, t+1\}} P (y_i | x_i, r^j_i)
    \end{equation}
    
    \item \textbf{On-policy Selector.} Similar with off-policy selector, but we use the \textit{Self-Reasoning Language Model} at last iteration to calculate the probability.
\end{itemize}

Afterwards, we can get the new generated instruction-tuning dataset $D^{t+1}$ = $\{(x_i, r^{t+1}_i, y_i)\}$ for next iteration, which completes the whole self-improved processing.

\section{Experiments}

\begin{figure*}
    \centering
     \includegraphics[width=1.0\textwidth]{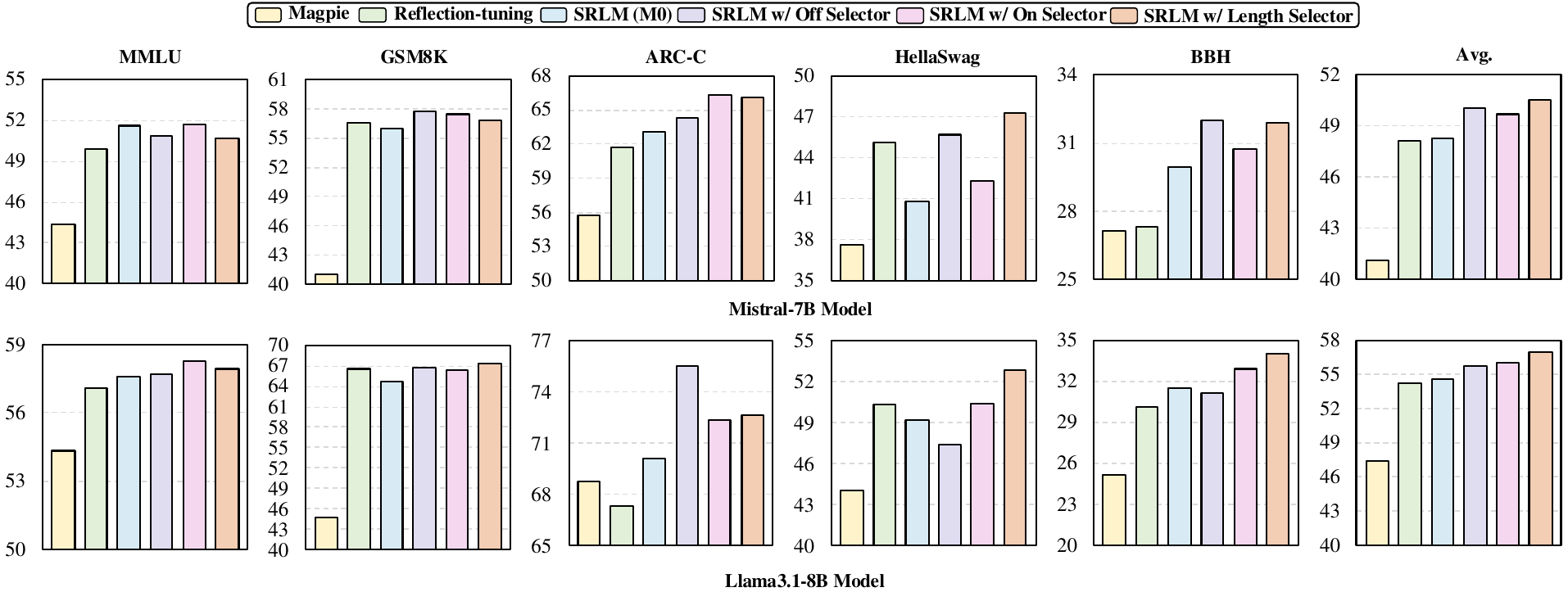}
    \caption{The performance of Self-Reasoning Language Models (SRLM) across five different benchmarks, along with the average performance (Avg.). We report the best performance during the five iterations for each selector, and the detailed experimental logs can be referred to at Table~\ref{tab:exp_log}. We also run significant test which showcases our method significantly outperforms the reflection-tuning with $p < 0.05$.}
    \label{fig:main_exp_results}
\end{figure*}

\subsection{Set Up}

\paragraph{Models.} We mainly use two different base models: Llama3.1-8B \citep{grattafiori2024llama3herdmodels} and Mistral-7B-V0.3 \citep{jiang2023mistral7b} since both of them are open-sourced and support longer context.

\paragraph{Data.} There are two different parts of data used in our experiments: 1) \textit{instruction-tuning data}: we first randomly sample 50k instances from Magpie-reasoning-150k \citep{xu2024magpiealignmentdatasynthesis} and then adopt reflection-tuning \citep{li-etal-2024-selective} to reflect its instruction and the responses using GPT-4o model; 2) \textit{reasoning catalyst data}: we use the same GPT-4o model\footnote{We use this model since our initial instruction-tuning data also is generated by this model.} to generate only additional 1k samples (only 0.02\% of whole training dataset) to enrich the original rationales with various meta-reasoning skills. We compare the performance of various models fine-tuned using our method against two above instruction-tuning datasets 50k Magpie data and 50k Reflection-tuning data (i.e., \textbf{Magpie} and \textbf{Reflect-tuning}).

\paragraph{Evaluation and Metrics.} We choose several existing benchmarks to validate the effectiveness and generalization of our proposed method. Specifically, we mainly focus on tasks necessniate complex reasoning such as general reasoning benchmarks: MMLU \citep{mmlu}, mathematical reasoning: GSM8K \cite{cobbe2021trainingverifierssolvemath}, commonsense reasoning: ARC-C \cite{clark2018thinksolvedquestionanswering}, HellaSwag \cite{zellers2019hellaswagmachinereallyfinish} and other challenging tasks: BBH \citep{suzgun2022challenging}. We choose Accuracy as the evaluation metric and use the zero-shot prompting method to evaluate the performance of fine-tuned models, following lots of previous works \cite{grattafiori2024llama3herdmodels}.

% \begin{itemize}
%     \item Magpie: Vanilla setting refers to using the original, unmodified IFT datasets for fine-tuning LLMs such as \textsc{Mistral} and \textsc{Llama-3}. This serves as a baseline to compare the effectiveness of dataset revision techniques.
% \end{itemize}

% \paragraph{SR} Sample Revision~(SR) marks the baseline for revising the instruction-answer pairs without leveraging any internal knowledge from the target LLM $\mathcal{M}$. This lets SR solely infuse knowledge from the revisor agent $\mathcal{A}_r$ into IFT datasets. 

\paragraph{Implementation Details.} We train 1,000 steps on 16 A100 80G GPUs for each method to ensure a fair comparison, considering the slightly different sizes of the dataset. During the training, we set the learning rate as 1e-05, batch size as 4, and gradient accumulation steps as 2 following \citet{zheng-etal-2024-llamafactory}. After obtaining the SRLM, we set the sample times $N$ in Eq~\ref{sample_eq} as 5, and the maximum iteration times as 5 for all selectors. During the inference, we adopt the same setting with temperature as 0.2, top p as 0.9, and max tokens as 8192. We denote the iterations of SRLM as $\mathcal{M}_{0,1,..,t}$.

\subsection{Main Results}
The main results are shown in Figure~\ref{fig:main_exp_results}. Based on that, we have the following key observations.

\paragraph{Incorporating additional reasoning expansion samples into the instruction-tuning dataset acts as a \textit{catalyst}, leading to improved performance.} As we can find in the Avg. and most benchmarks, SRLM ($\mathcal{M}_0$) achieves better performance than Reflection-tuning and Magpie in both models. The only two exceptions are GSM8K and HellaSwag, due to the relatively smaller number of instances of mathematical and commonsense reasoning in the reasoning expansion data.

\paragraph{Small Self-Reasoning Language Models can generate high-quality rationales iteratively.} It is evident that SRLM with different selectors at various iterations all leads to improvement compared with the initial SRLM ($\mathcal{M}_0$). Since the only difference between these SRLMs is the used instruction-tuning dataset, it strongly demonstrates that the SRLM with a small size (7B or 8B) can generate better instruction-tuning samples than GPT-4o (i.e., the instruction-tuning data at the $0_{th}$ step). It is encouraging to find that synthesized data by small language models still can provide stronger supervision signals than the latest powerful models. In addition, we find that Llama3.1-8B achieves more consistent and higher improvement than Mistral-7B since it generates better samples due to more strong base capabilities.

\paragraph{All types of selectors outperform SRLM ($\mathcal{M}_0$), with the on-policy selector excelling in MMLU, while the length selector performs better on HellaSwag and BBH.} Generally, Length selector achieves the best performance at Avg. for both models, revealing the excellence of this simple strategy in SRLM. In detail, we observe a seesaw phenomenon across different datasets and models. For example, with the same Mistral-7B model, the on-policy selector performs better on ARC-C but worse on HellaSwag. However, the on-policy selector on the Llama3.1-8B model performs worse compared to the other selectors on ARC-C. This can be attributed to the interaction effects between the models and datasets, highlighting the challenges in evaluating self-improving LLMs.

\begin{figure*}
    \centering
    \includegraphics[width=0.9\textwidth]{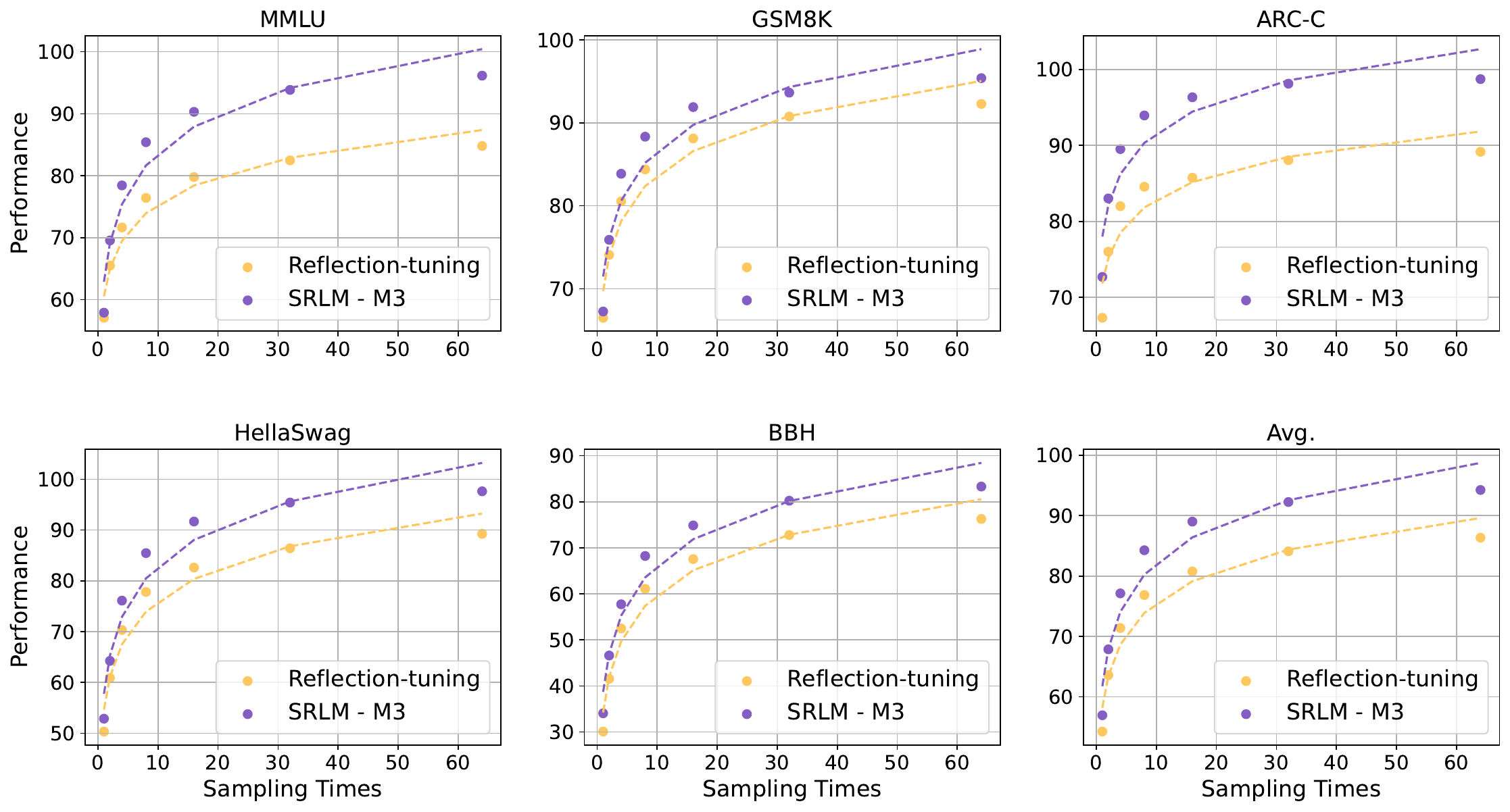}
    \caption{The performance of the baseline and our proposed SRLM ($\mathcal M_3$ with length selector) different sampling times ranging from 1 to 64 (i.e., 1, 2, 4, 8, 16, 32, 64) on five various benchmarks and the Avg. performance.}
    \label{fig:best-of-n}
\end{figure*}

\section{Analysis}

\subsection{Effects of Sampling Times.}

In the main experiments, we mainly consider the best-of-1 performance of different methods. Alternatively, the best-of-N performance has been proved to be an effective way to showcase the depth and creativity of the model's problem-solving potential in terms of reasoning \cite{chow2024inferenceaware}. In this way, we conduct the multiple sampling ranging from 1 to 64, aiming to investigate the performance under the best-of-N setting. Figure~\ref{fig:best-of-n} shows the performance of our proposed SRLM ($\mathcal{M}_3$ with length selector) and the baseline. The full experimental results can be found in Table~\ref{tab:sampling_times} and fitting functions can be found in Appendix. First of all, it is worth noting that SRLM achieves much better performance with more sampling times on all benchmarks, as the gains get bigger and bigger. This is further supported by the larger coefficient of the logarithmic term in the fitted function of SRLM. Secondly, we can observe that MMLU achieves the highest growth rate and then followed by ARC-C and HellaSwag. As shown in Table~\ref{tab:sampling_times}, SRLM achieves 94.24\% on the Avg. performance with 64 sampling times with MMLU (96.15\% ($+11.35$)), ARC-C (98.72\% ($+9.56$)), HellaSwag (97.62\% ($+8.38$)). Based on above observations, it is believed that our proposed SRLM can explore more depth, diverse and creative reasoning paths toward the final correct solution, revealing its promising potential.

\subsection{Effects of Different Number of Iterations}

\begin{figure}
    \centering
    \includegraphics[width=0.49\textwidth]{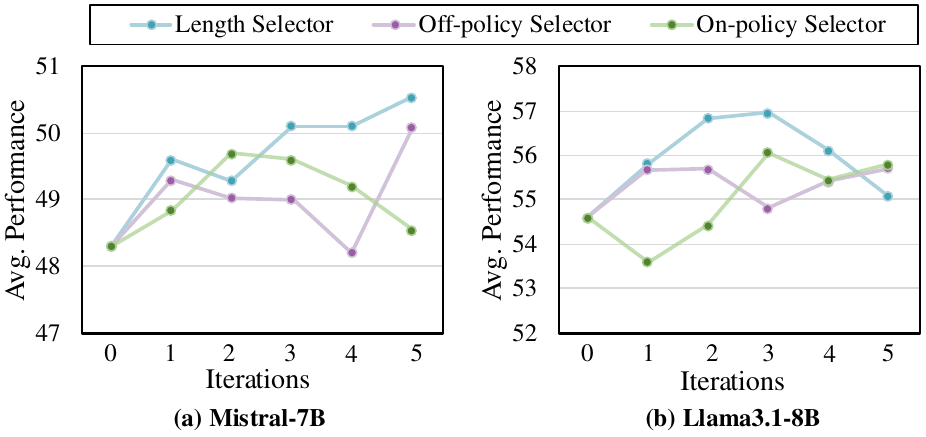}
    \caption{The effects of different number of iterations in Self-Reasoning Language Models.}
    \label{fig:iterations_effects}
\end{figure}

Figure~\ref{fig:iterations_effects} presents the Avg. Performance under different iterations with three types of selectors. There are several key findings. On the one hand, on-policy methods typically perform poorly at the beginning of the iteration process. This can be attributed to the limited number of samples - only a few hundred - selected during the initial iteration. Moreover, as the process progresses from the first to the third iteration, the number of selected samples increases with the upward trend of the performance observed for both models in the figure. After the third iteration, performance diverges: the number of selected samples decreases for Mistral-7B, leading to a performance decline, while it remains constant for Llama3.1-8B, resulting in stable and saturated performance.

On the other hand, the length selector mostly leads to the best performance of the three selectors, and we also observe model degradation no matter which selector is chosen. After checking the number of selected numbers at each iteration, we find that the off-policy selector selects the most samples in the first iteration, nearly matching the original dataset size, but drops sharply to around 10k and continues decreasing. In contrast, the length-based selector also decreases, but at a slower and steadier rate. The last thing we want to emphasize is that it is evident that SRLM is capable of correcting previous rationales considering that there may be errors in the enriched rationales in previous steps, as we can find that length and off-policy selector achieve the best performance at $5_{th}$ iteration with the worst performance at $4_{th}$.

% since the performance Llama3.1-8B 

% recover from noise rationales, seesaw phenomenon iterations

\subsection{Effects of Reasoning Catalyst Data}

\begin{figure}
    \centering
\includegraphics[width=0.49\textwidth]{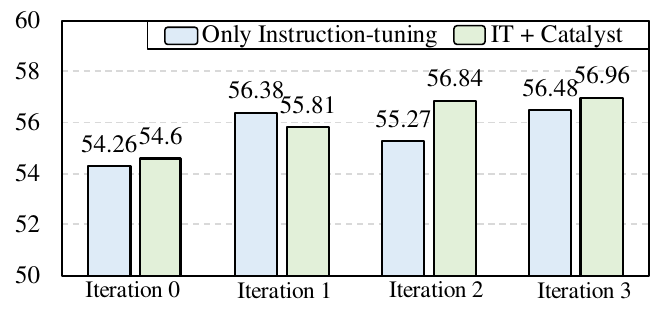}
    \caption{The ablation study by removing the reasoning catalyst data at each iteration. We report the Avg. performance here and the detailed results can be found in Table~\ref{tab:catalyst_ablation}.}
    \label{fig:ablation_study}
\end{figure}

We also investigate several factors of reasoning catalyst data that may have serious effects on the final performance of SRLM, including the ablation study, different sizes, and sources of the data.

\paragraph{Ablation Study.} We first directly remove the catalyst data from each iteration and train the model independently following the same setting in the main experiments. Figure~\ref{fig:ablation_study} presents the final results at different iterations. Two obvious observations can be drawn from the results. First of all, the performance at later iterations mostly outperforms the performance at the initial iteration (iteration at $t=0$) no matter incorporates a reasoning catalyst or not. This provides additional direct evidence that the newly generated instruction-tuning datasets at iterations ($t>1$) are better than the GPT-4o generated dataset at iteration $t=0$. Secondly, when incorporating reasoning catalyst data, the improvements become more stable and consistent. We can find that IT + Catalyst mostly achieves better performance than only instruction-tuning data, and it brings consistent improvement while the performance using instruction-tuning data only may fluctuate.

\begin{table}[!t]
\setlength{\belowcaptionskip}{0pt}
    \centering
    \begin{adjustbox}{max width=0.49 \textwidth}
    \begin{tabular}{l ccccc|c}
    \toprule
    \textbf{Data}  & MMLU & GSM8K & ARC-C & HellaSwag & BBH & Avg. \\
    \hline
    Reflection-tuning & 57.08 & 66.49 & 67.32 & 50.34 & 30.09 & 54.26 \\

    \hline

    SRLM -- $\mathcal{M}_0$ \\
    \hdashline

    \quad Cat. = 1k & 57.60 & 64.59 & \textbf{70.14} & \textbf{49.23} & 31.46 & \textbf{54.60} \\
    \quad Cat. = 5k & 57.46 & 66.26 & 67.15 & 47.87 & 25.23 & 52.79 \\
    \quad Cat. = 10k & \textbf{58.81} & \textbf{67.63} & 67.41 & 46.30 & \textbf{32.29} & 54.49 \\

    SRLM -- $\mathcal{M}_1$ \\
    \hdashline
    
    \quad Cat. = 1k & 58.31 & 65.73 & 72.87 & \textbf{48.05} & \textbf{34.08} & 55.81 \\
    \quad Cat. = 5k & 58.03 & \textbf{67.85} & \textbf{74.23} & 46.54 & 32.51 & \textbf{55.83} \\
    \quad Cat. = 10k &  \textbf{58.55} & 67.63 & 73.29 & 45.27 & 31.78 & 55.30 \\

    SRLM -- $\mathcal{M}_2$ \\
    \hdashline

    \quad Cat. = 1k & 58.27 & \textbf{69.22} & 72.70 & \textbf{51.99} & 32.03 & \textbf{56.84} \\
    \quad Cat. = 5k & 56.96 & 67.17 & 71.16 & 47.11 & 28.58 & 54.20 \\
    \quad Cat. = 10k & \textbf{58.34} & 67.40 & \textbf{73.38} & 51.06 & \textbf{33.18} & 56.67 \\

    SRLM -- $\mathcal{M}_3$ \\
    \hdashline
    
    \quad Cat. = 1k &  57.91 & \textbf{67.25} & \textbf{72.70} & \textbf{52.88} & \textbf{34.06} & \textbf{56.96} \\
    \quad Cat. = 5k & 57.30 & 65.73 & 70.73 & 46.97 & 30.61 & 54.27 \\
    \quad Cat. = 10k & \textbf{58.24} & 67.17 & 70.56 & 50.87 & 32.83 & 55.93 \\
    
    \bottomrule
    \end{tabular}
    \end{adjustbox}
    \caption{The results using different size of reasoning catalyst data based on Llama3.1-8B model with 1,000 steps optimization on three iterations ($\mathcal{M}_{0,1,2,3}$). We \textbf{bold} the best performance at each iteration.}
    \label{tab:catalyst_effects}
\end{table}

\begin{table}[!t]
\setlength{\belowcaptionskip}{0pt}
    \centering
    \begin{adjustbox}{max width=0.48 \textwidth}
    \begin{tabular}{l cccccc}
    \toprule
    \textbf{Data}  & MMLU & GSM8K & ARC-C & HellaSwag & BBH & Avg. \\
    \hline
    Reflection-tuning & 57.08 & 66.49 & 67.32 & 50.34 & 30.09 & 54.26 \\
    \hdashline
    IT$_0$ + New Cat & 57.69 & 66.34 & 70.90 & 49.32 & 32.22 & 55.30 \\
    IT$_1$ + New Cat & 56.77 & 68.31 & 69.28 & 45.53 & 33.40 & 54.66 \\
    IT$_2$ + New Cat & 58.17 & 67.40 & 70.31 & 48.10 & 31.28 & 55.05 \\
    IT$_3$ + New Cat & 56.45 & 66.49 & 69.62 & 46.47 & 30.27 & 53.86 \\
    \bottomrule
    \end{tabular}
    \end{adjustbox}
    \caption{The performance of SRLM with length selector on Llama3.1-8B using \textit{New} 1k reasoning catalyst data with 1000 steps optimization.}
    \label{tab:new_catalyst}
\end{table}

\paragraph{Size of Catalyst.} We additionally random sample 5,000 and 10,000 from the original instruction-tuning dataset and build corresponding catalyst data. We also train the SRLM with 1,000 optimization steps to make a fair comparison following the same setting in the main experiments. Table~\ref{tab:catalyst_effects} shows the final results. Generally, adding more reasoning catalyst data does not significantly improve average performance. However, it can lead to noticeable gains on specific datasets.

In detail, SRLM ($\mathcal{M}_0$) with 10,000 reasoning catalyst data achieves best performance at MMLU, GSM8K and BBH. Moreover, it continues to show superior results on MMLU with additional training iterations. This evidence suggests that incorporating more catalyst reasoning data benefits general reasoning tasks that require comprehensive evaluation, rather than domain-specific reasoning tasks. Furthermore, we can find that it offers limited benefits for commonsense reasoning tasks like ARC-C and HellaSwag, as the SRLM with just 1,000 data points consistently achieves peak performance across all iterations. We attribute this to relatively smaller ratio of commonsense reasoning samples in the sampled dataset.

\begin{table}[!t]
\setlength{\belowcaptionskip}{0pt}
    \centering
    \begin{adjustbox}{max width=0.4 \textwidth}
    \begin{tabular}{l cc}
    \toprule
    \textbf{Method}  & Arena-Hard & Alpaca Eval V2 \\
    \hline
    Alpaca-7B \hspace{20mm} & - & 2.59 \\
    Reflection-tuning & 6.40 & 5.92 \\
    \hdashline
    SRLM (\textit{Ours}) & 8.60 & 9.21 \\
    \bottomrule
    \end{tabular}
    \end{adjustbox}
    \caption{The win rate of SRLM and Reflection-tuning}
    \label{tab:alpaca_eval}
\end{table}

\paragraph{New Catalyst.} We also investigate the effects of applying out-of-distribution reasoning catalysts. To achieve this, we construct the catalyst using a separate randomly sampled 1,000 instruction-tuning dataset that has no overlap with the existing datasets. Table~\ref{tab:new_catalyst} shows the performance at each iteration. It is shown that incorporating additional out-of-distribution reasoning catalyst data also brings improvement compared with the only 4o-generated instruction-tuning data. However, consistent improvement in later iterations is not guaranteed as we observe only partial gains in certain benchmarks rather than a significant increase in overall average performance. We consider this due to difficulty in refining the rationale for existing instructions given out-of-distribution demonstrations.

\paragraph{Update Catalyst.} Until now, we only update the instruction-tuning data set as shown in Figure~\ref{fig:ggl}. Alternatively, we can also update reasoning catalyst data by utilizing the newly generated rationales $r_j^t$. In detail, we can update the $r_j^*$ in Figure~\ref{fig:ggl} only with $r_j^{t+1}$ \textit{or} update $r_j^0$ and $r_j^*$ with $r_j^{t}$ and $r_j^{t+1}$ at the same time with simple length selector. However, we find that both of these two updating mechanisms can not bring significant improvement and even downgrade the performance at the later iterations as shown in Table~\ref{tab:update_catalyst}. We consider the primary reason may be that LLMs tend to become constrained by their own outputs, limiting exploration of diverse solutions when more self-generated rationales are included. Moreover, the reasoning catalyst data introduced at the initial stage may be nearly optimal, thus being difficult to beat with only length selector. More analysis can be found in Appendix.

\section{Discussion}

\paragraph{LLM-as-a-Judge Evaluation.} Besides the exact match evaluation of five various reasoning benchmarks, we also conduct LLM-as-a-judge evaluation on two  popular benchmarks: Arena-Hard \cite{arena-hard} and Alpaca Eval V2 \cite{alpaca_eval}. Following the default setting, we utilize gpt-4-1106-preview to compare the responses generated by different methods (i.e., Reflection-tuning and SRLM - $\mathcal{M}_3$ with length selector) against the response generated by the reference model (i.e., gpt-4-0314 in Arena-Hard, gpt-4-turbo in Alpaca Eval). Table~\ref{tab:alpaca_eval} shows the win rate of response pairs. It is clear that our proposed SRLM achieves higher win rates at both datasets, leading to relatively 34\% and 56\% improvement on Arena-Hard and Alpaca Eval V2 respectively.

\paragraph{Effects of Instruction-tuning Data.}

\begin{table}[!t]
\setlength{\belowcaptionskip}{0pt}
    \centering
    \begin{adjustbox}{max width=0.48 \textwidth}
    \begin{tabular}{l ccccc|c}
    \toprule
    \textbf{Data}  & MMLU & GSM8K & ARC-C & HellaSwag & BBH & Avg. \\
    \hline
    Reflection-tuning & 51.44 & 48.07 & 66.64 & 37.24 & 31.71 & 47.02 \\
    \hline
    SRLM (\textit{Ours}) \\
    \hdashline
    \quad IT$_0$ + Cat. & \textbf{55.34} & 49.81 & \textbf{69.28} & 42.88 & 30.57 & 49.58 \\
    \quad IT$_1$ + Cat. & 53.85 & \textbf{51.78} & 67.58 & 44.93 & \textbf{34.07} & \textbf{50.44} \\
    \quad IT$_2$ + Cat. & 52.81 & 49.81 & 68.17 & \textbf{47.35} & 31.21 & 49.87 \\
    \quad IT$_3$ + Cat. & 52.31 & 47.08 & 66.47 & 45.03 & 28.53 & 47.89 \\
    \bottomrule
    \end{tabular}
    \end{adjustbox}
    \caption{The results of Llama3.1-8B on instruction-tuning dataset (collected by length selector) of each iteration for Alpaca \cite{alpaca} dataset with 2,000 optimization steps.}
    \label{tab:alpaca_res}
\end{table}

To further validate the robustness of our proposed SRLM, we also conduct our experiments on another popular instruction-tuning dataset: Alpaca \cite{alpaca}. Following the same procedure, we finetuned our models for all datasets in 2,000 steps with batch size at 2 due to more longer responses in original Alpaca dataset. Table~\ref{tab:alpaca_res} shows the final results. It is clear that our method still outperforms the competitive 4o-generated data by reflection-tuning, and achieves best performance at the first iteration with the absolute $+3.42$ improvement on average performance of five different benchmarks. This result evidences the robustness and generalization of our proposed method on different datasets.

\begin{figure}
    \centering
    \includegraphics[trim={0cm 0cm 0cm 0cm}, clip, width=0.5\textwidth]{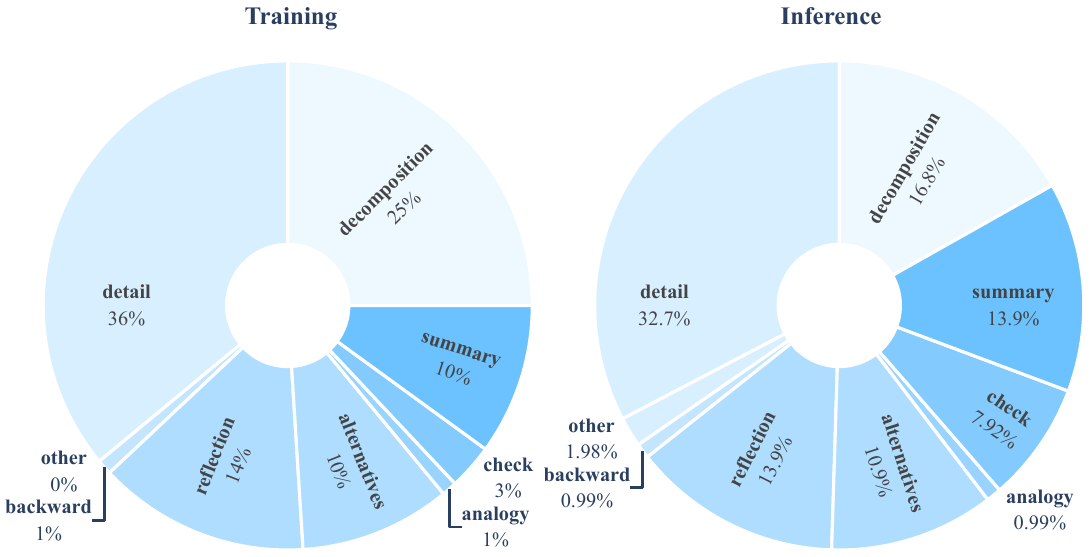}
    \caption{The distribution of various meta-reasoning skills employed at the training and inference of SRLM.}
    \label{fig:tag_distribution}
\end{figure}

\paragraph{Meta-Reasoning Skills Distribution.} We also study the distribution of various meta-reasoning skills employed during training and inference stage of SRLM. Figure~\ref{fig:tag_distribution} shows the results. On the one hand, it is obvious that \textit{detail} is the dominant reasoning skills used to unfold hidden reasoning rationales, and then followed by \textit{decomposition}, \textit{reflection}, no matter in training and inference stage. On the other hand, the striking similarity in reasoning skill distribution patterns between training and inference phases strongly suggests that instruction-tuning data construction critically influences model behavior. The distribution of various reasoning skills across different benchmarks during inferences also shows similar trends (as shown in Figure~\ref{fig:all_tags_inference}).

\section{Conclusion}

In this paper, we present Self-Reasoning Language Models (SRLM), which is capable to generate responses and unfold hidden reasoning chains at the same time, leading to self-evolved models that enhance performance across various reasoning benchmarks. While there are many avenues left unexplored, we believe this is exciting and promising since the SRLM is better able to discover diverse and creative reasoning paths in future iterations for improving instruction following – a kind of virtuous circle.

% SRLM can enrich previous reasoning rationales with the augmentation of many meta-reasoning skills in each iteration, i.e., \textit{detail}, \textit{decomposition} and \textit{reflection}. Therefore, 

% By incorporating additional reasoning catalyst data with original instruction-tuning dataset to learn how to reason, SRLM can iteratively refine its own rationales and evolves with better instruction-tuning dataset at each iteration. 

\section*{Limitations}

Despite SRLM demonstrates effectiveness acorss various models and benchmarks, due to the limitation of computation resources, we do not test SRLM on more larger LLMs such as 70B or even larger models. In addition, SRLM can be applied to datasets or benchmarks with verifiable answers and trained continuously, as it is independent of the data used and training methods. We leave these aspects for future work, as they are beyond the scope of this paper.

% \paragraph{More .} The performance of SRLM highly related with the used reasoning catalyst data. We mainly leverage GPT-4o to generate the data while recent open-source models such as QwQ and Deepseek-R1 already produce o1-like reasoning paths. It is worthy exploring more effective reasoning catalyst data in future work.

% \paragraph{Training Method.} We mainly train SRLM from scratch aiming to directly showcase the quality of collected dataset. However, we consider there may exist other training methods, such as continual train the tuned SRLM in previous iteration, may lead to better performance.

% 1) error reasoning rationales introduced
% 2) catalyst comes from superior model o1-like models
% 3) continual training

\section*{Ethical Considerations}

In conducting our research, we have thoroughly reviewed and ensured compliance with ethical standards. Our study utilizes existing datasets, which have been publicly available and previously vetted for ethical use. These datasets have been carefully selected to avoid any form of offensive or biased content. Therefore, we consider that our research does not present any ethical issues.

\section*{Acknowledgments}

This work was partially supported by Hong Kong RGC GRF No. 14206324, CUHK direct grant No. 4055209, and CUHK Knowledge Transfer Project Fund No. KPF23GWP20.

\bibliography{custom}

\appendix

\clearpage

\section{Appendix}
\label{sec:appendix}

\subsection{Prompt Details}

Table~\ref{table:reasoning_generation_prompt} and Table~\ref{table:meta_reasoning_prompt} show the instruction and system prompt respectively used to instruct LLMs to unfold the hidden reasoning chains with the augmentation of various meta-reasoning skills.

\begin{table}[ht]
\small
    \centering
    \colorbox{blue!8}{
    \begin{tabular}{@{}p{7.2cm}}
    Here is the given question: \{ins\} \\\\
    
    Here is the original reasoning: \{ans\} \\\\
    
    \end{tabular}
    }
    \caption{The instruction used to prompt LLM to generate the enriched rationales.}
    \label{table:reasoning_generation_prompt}
\end{table}

\begin{table*}[ht]
\small
    \centering
    \colorbox{blue!8}{
    \begin{tabular}{@{}p{16.2cm}}
    You are an expert at meta reasoning theory from cognitive science. Given the question, corresponding summarized reasoning and answer, you can always uncover hidden or unspoken reasoning, even when it isn't explicitly stated.

    You need to add any missing reasoning thoughts that you think it is helpful or may occurs to understand and solve this question based on given summarized reasoning result. Your new reasoning should be more comprehensive, detailed and clear. Your answer should follow the format like <thoughts> your new reasoning here </thoughts>
    
    Inside <thoughts> </thoughts>, you need to explicitly indicate which meta reasoning skill is used, some of them are shown below, but note you can use anything else if you think it is helpful or it naturally occurs when you solve this question. \\\\
    
    <decomposition>: breaking down a complex problem into smaller, more manageable parts. Making sure that you also provide answers for all decomposed problems in this section. You can decompose iterativelly but should not contain same problem or exceed the max iteration depth which is three. \\\\
    
    <backward>: starting with the desired observations at any previous reasoning step and working backward to identify the new reasoning directions. \\\\
    
    <detail>: any details including but not limited to logic and reasons for your reasoning in this way, you are encouraged to add this at every unclear or unnatural reasoning step. \\\\
    
    <summary>: summarize your reasoning to help future thinking. \\\\
    
    <alternatives>: directly thinking in other ways, try to explore different solutions as much as possible to solve given problem. \\\\
    
    <reflection>: you are encouraged to regularly reflect on your past reasoning in current response at various levels of detail, from sentence down to individual word. This will help you better understand and think through problems. It's okay to make mistakes; use them as opportunities to learn and improve. \\\\
     
    <analogy>: you are encouraged to regularly consider other analogous problems with the problem you've encountered at various reasoning steps, along with their solutions. Reference existing theories or methods that guided your approach to solving these problems. These similar problems can be at various levels of detail - from larger overarching issues down to smaller sub-problems you encountered along the way. The key is to demonstrate a diverse range of problems and solutions, to show how you have approached and resolved challenges that are analogous to the current situation. \\\\
    
    <check>: consider different edge cases or test cases carefully. \\\\

    <other>: other meta reasoning skills you think is helpful or worthy to try to solve the task. \\\\
    
    Notice: \\\\

    1. All tags must be properly invoked and closed, using the format like <reflection> and </reflection>. \\\\
    2. You should always use first-person perspective. \\\\
    3. You can add any new meta reasoning skill at any positions except <reflection>. Note <reflection> can not be invoked without any reasoning in current response and it can be invoked at any positions when you already have some reasoning results. \\\\
    4. You cannot change the original reasoning. However, if you identify any errors or improvements in the reasoning, you can add new reasoning steps using above meta-reasoning skills afterwards to correct or clarify the path, ensuring a better understanding and solution. \\\\
    5. You can apply the same reasoning skills multiple times or use different skills simultaneously. \\\\
    6. Your answer should start with <thoughts>, and end with </thoughts>. \\\\

    \end{tabular}
    }
    \caption{The system prompt used to initialize LLM to unfold all hidden reasoning chains.}
    \label{table:meta_reasoning_prompt}
\end{table*}

\subsection{Experimental Logs}

\paragraph{Reflection-tuning.} 
MMLU: $y = 6.4535 * log(x) + 60.5418$;
GSM8K: $y = 6.0944 * log(x) + 69.6957$;
ARC-C: $y = 4.8088 * log(x) + 71.8375$;
HellaSwag: $y = 9.2765 * log(x) + 54.6557$;
BBH: $y = 11.1345 * log(x) + 34.2507$;
Avg.: $y = 7.5551 * log(x) + 58.1925$

\paragraph{SRLM.}
MMLU: $y = 9.0256 * log(x) + 62.8932$
GSM8K: $y = 6.5905 * log(x) + 71.4682$
ARC-C: $y = 5.9295 * log(x) + 78.0043$
HellaSwag: $y = 10.9284 * log(x) + 57.7607$
BBH: $y = 11.9599 * log(x) + 38.7086$
Avg.: $y = 8.8870 * log(x) + 61.7671$

\subsection{Performance on Qwen-14B}

We also conduct experiments on Qwen-14B models following the same setting in main experiments. Table~\ref{tab:more_models} shows the final results. It is observed that SRLM initially exhibits higher performance at the 0th iteration but then gradually degrades. We attribute this to the ineffective of reasoning catalyst data to guide the reasoning processing since Qwen2.5-14B already achieves comparable performance against the used 4o models to generate reasoning catalyst data.

\begin{table*}[!t]
\setlength{\belowcaptionskip}{0pt}
    \centering
    \begin{adjustbox}{max width=0.96 \textwidth}
    \begin{tabular}{llccccc|c}
    \toprule
    \textbf{Models} & \textbf{Methods}  & MMLU & GSM8K & ARC-C & HellaSwag & BBH & Avg. \\
    \hline

    \multirow{15}{*}{\textbf{Mistral-7B}} 
    & Magpie & 44.35 & 41.02 & 55.72 & 37.61 & 27.13 & 41.17 \\
    & Reflection-tuning & 49.85 & 56.56 & 61.69 & 45.15 & 27.29 & 48.11 \\

    \cline{2-8}
    
    & \quad IT$_{0}$ + Cat. ($\mathcal{M}_0$) & 51.63 & 56.03 & 63.05 & 40.82 & 29.97 & 48.30 \\

    & \textit{Length Selector} \\
    \cdashline{2-8} 
    & \quad IT$_{1}$ + Cat. ($\mathcal{M}_1$) & 49.50 & 58.91 & 62.37 & 45.91 & 31.33 & 49.60 \\
    & \quad IT$_{2}$ + Cat. ($\mathcal{M}_2$) & 50.01 & 59.59 & 64.85 & 42.16 & 29.82 & 49.29 \\
    & \quad IT$_{3}$ + Cat. ($\mathcal{M}_3$) & 50.56 & 56.41 & 64.59 & 46.88 & 32.11 & 50.11 \\
    & \quad IT$_{4}$ + Cat. ($\mathcal{M}_4$) & 49.07 & 58.45 & 60.41 & 42.03 & 29.35 & 47.86 \\
    & \quad IT$_{5}$ + Cat. ($\mathcal{M}_5$) & 50.66 & 56.79 & 66.13 & 47.30 & 31.87 & 50.55 \\

    & \textit{Off-policy Selector} \\
    \cdashline{2-8} 
    & \quad IT$_{1}$ + Cat. ($\mathcal{M}_1$) & 49.55 & 58.07 & 65.02 & 41.02 & 32.82 & 49.30 \\
    & \quad IT$_{2}$ + Cat. ($\mathcal{M}_2$) & 50.63 & 57.92 & 62.12 & 41.98 & 32.49 & 49.03 \\
    & \quad IT$_{3}$ + Cat. ($\mathcal{M}_3$) & 49.74 & 55.95 & 65.02 & 42.28 & 32.12 & 49.02 \\
    & \quad IT$_{4}$ + Cat. ($\mathcal{M}_4$) & 50.19 & 56.56 & 60.24 & 42.55 & 31.48 & 48.20 \\
    & \quad IT$_{5}$ + Cat. ($\mathcal{M}_5$) & 50.79 & 57.77 & 64.25 & 45.66 & 32.00 & 50.09 \\

    & \textit{On-policy Selector} \\
    \cdashline{2-8} 
    & \quad IT$_{1}$ + Cat. ($\mathcal{M}_1$) & 50.00 & 58.68 & 64.59 & 41.27 & 29.64 & 48.84 \\
    & \quad IT$_{2}$ + Cat. ($\mathcal{M}_2$) & 51.69 & 57.47 & 66.30 & 42.29 & 30.77 & 49.70 \\
    & \quad IT$_{3}$ + Cat. ($\mathcal{M}_3$) & 50.28 & 56.33 & 64.59 & 46.44 & 30.36 & 49.60 \\
    & \quad IT$_{4}$ + Cat. ($\mathcal{M}_4$) & 51.66 & 55.34 & 65.36 & 44.55 & 29.07 & 49.20 \\
    & \quad IT$_{5}$ + Cat. ($\mathcal{M}_5$) & 49.54 & 55.57 & 62.63 & 43.00 & 32.00 & 48.55 \\
    
    \hline
    
    \multirow{15}{*}{\textbf{Llama3.1-8B}}
    & Magpie & 54.34 & 44.81 & 68.77 & 44.05 & 25.10 & 47.41 \\
    & Reflection-tuning & 57.08 & 66.49 & 67.32 & 50.34 & 30.09 & 54.26 \\

    \cline{2-8}
    
    & \quad IT$_{0}$ + Cat. ($\mathcal{M}_0$) & 57.60 & 64.59 & 70.14 & 49.23 & 31.46 & 54.60 \\
    
    & \textit{Length Selector} \\
    \cdashline{2-8} 
    & \quad IT$_{1}$ + Cat. ($\mathcal{M}_1$) & 58.31 & 65.73 & 72.87 & 48.05 & 34.08 & 55.81\\
    & \quad IT$_{2}$ + Cat. ($\mathcal{M}_2$) & 58.27 & 69.22 & 72.70 & 51.99 & 32.03 & 56.84 \\
    & \quad IT$_{3}$ + Cat. ($\mathcal{M}_3$) & 57.91 & 67.25 & 72.70 & 52.88 & 34.06 & 56.96 \\
    & \quad IT$_{4}$ + Cat. ($\mathcal{M}_4$) & 57.27 & 67.17 & 72.87 & 49.83 & 33.44 & 56.12 \\
    & \quad IT$_{5}$ + Cat. ($\mathcal{M}_5$) & 56.17 & 65.73 & 69.45 & 50.27 & 33.75 & 55.07 \\

    & \textit{Off-policy Selector} \\
    \cdashline{2-8} 
    & \quad IT$_{1}$ + Cat. ($\mathcal{M}_1$) & 58.26 & 67.70 & 74.23 & 46.11 & 32.14 & 55.69 \\
    & \quad IT$_{2}$ + Cat. ($\mathcal{M}_2$) & 58.50 & 67.25 & 73.46 & 48.03 & 31.28 & 55.70 \\
    & \quad IT$_{3}$ + Cat. ($\mathcal{M}_3$) & 58.30 & 65.43 & 73.04 & 46.93 & 30.37 & 54.81 \\
     & \quad IT$_{4}$ + Cat. ($\mathcal{M}_4$) & 58.57 & 67.40 & 74.15 & 46.93 & 30.03 & 55.42 \\
    & \quad IT$_{5}$ + Cat. ($\mathcal{M}_5$) & 57.72 & 66.79 & 75.51 & 47.40 & 31.12 & 55.71 \\

    & \textit{On-policy Selector} \\
    \cdashline{2-8} 
    & \quad IT$_{1}$ + Cat. ($\mathcal{M}_1$) & 57.96 & 65.20 & 70.05 & 45.07 & 29.69 & 53.59 \\
    & \quad IT$_{2}$ + Cat. ($\mathcal{M}_2$) & 58.53 & 64.67 & 69.20 & 48.41 & 31.32 & 54.43 \\
    & \quad IT$_{3}$ + Cat. ($\mathcal{M}_3$) & 58.26 & 66.41 & 72.35 & 50.43 & 32.89 & 56.07 \\
    & \quad IT$_{4}$ + Cat. ($\mathcal{M}_4$) & 58.45 & 68.76 & 70.14 & 46.54 & 33.37 & 55.45 \\
    & \quad IT$_{5}$ + Cat. ($\mathcal{M}_5$) & 59.17 & 66.19 & 69.88 & 50.76 & 33.01 & 55.80 \\
    
    \bottomrule
    \end{tabular}
    \end{adjustbox}
    \caption{The detailed experimental results of each iteration.}
    \label{tab:exp_log}
\end{table*}

\begin{table*}[h]
\setlength{\belowcaptionskip}{0pt}
    \centering
    \begin{adjustbox}{max width=0.96 \textwidth}
    \begin{tabular}{llccccc|c}
    \toprule
    \textbf{Models} & \textbf{Methods}  & MMLU & GSM8K & ARC-C & HellaSwag & BBH & Avg. \\
    \hline

    \multirow{6}{*}{\textbf{Qwen2.5-14B}}
    & Reflection-tuning & 78.72 & 92.41 & \textbf{93.51} & 71.86 & 57.56 & 78.81 \\

    \cline{2-8}
    
    & \quad IT$_{0}$ + Cat. ($\mathcal{M}_0$) & 77.89 & 92.64 & 93.17 & \textbf{74.24} & \textbf{59.69} & \textbf{79.53} \\
    
    & \textit{Length Selector} \\
    \cdashline{2-8} 
    & \quad IT$_{1}$ + Cat. ($\mathcal{M}_1$) & 77.30 & \textbf{92.66} & 91.72 & 68.44 & 58.47 & 77.72 \\
    & \quad IT$_{2}$ + Cat. ($\mathcal{M}_2$) & 76.97 & 92.34 & 92.32 & 67..65 & 57.08 & 77.65 \\
    
    & \quad IT$_{3}$ + Cat. ($\mathcal{M}_3$) & 76.28 & 92.04 & 90.44 & 65.70 & 58.13 & 76.52 \\

    \bottomrule
    \end{tabular}
    \end{adjustbox}
    \caption{The detailed experimental results of each iteration on Qwen-14B with length selector.}
    \label{tab:more_models}
\end{table*}

\begin{table*}[!t]
\setlength{\belowcaptionskip}{0pt}
    \centering
    \begin{adjustbox}{max width=0.8 \textwidth}
    \begin{tabular}{l ccccc|c}
    \toprule
    \textbf{Data}  & MMLU & GSM8K & ARC-C & HellaSwag & BBH & Avg. \\
    \hline
    Reflection-tuning & 57.08 & 66.49 & 67.32 & 50.34 & 30.09 & 54.26 \\
    IT$_1$ & 57.74 & 67.93 & 74.06 & 49.22 & 32.94 & 56.38 \\
    IT$_2$ & 57.87 & 66.87 & 72.53 & 48.63 & 30.46 & 55.27 \\
    IT$_3$ & 57.51 & 68.46 & 73.21 & 51.59 & 31.61 & 56.48 \\
    \bottomrule
    \end{tabular}
    \end{adjustbox}
    \caption{The results of Llama3.1-8B on instruction-tuning dataset (collected by length selector) of each iteration.}
    \label{tab:catalyst_ablation}
\end{table*}

\begin{table*}[!t]
\setlength{\belowcaptionskip}{0pt}
    \centering
    \begin{adjustbox}{max width=0.8 \textwidth}
    \begin{tabular}{l ccccc|c}
    \toprule
    \textbf{Data}  & MMLU & GSM8K & ARC-C & HellaSwag & BBH & Avg. \\
    \hline
    % Reflection-tuning & \\

    SRLM -- $\mathcal{M}_0$ \\
    \hdashline

    \quad Cat. = 5k & 58.47 & 67.10 & 69.54 & 47.13 & 27.98 & 54.04 \\
    \quad Cat. = 10k &  59.22 & 66.26 & 67.58 & 46.68 & 30.78 & 54.10 \\

    SRLM -- $\mathcal{M}_1$ \\
    \hdashline
    
    \quad Cat. = 5k & 58.04 & 66.41 & 75.17 & 45.71 & 31.47 & 55.36 \\
    \quad Cat. = 10k & 58.56 & 66.72 & 75.00 & 48.00 & 30.02 & 55.66 \\

    SRLM -- $\mathcal{M}_2$ \\
    \hdashline

    \quad Cat. = 5k & 58.05 & 68.46 & 72.70 & 44.76 & 29.28 & 54.65 \\
    \quad Cat. = 10k & 57.84 & 67.70 & 73.04 & 50.79 & 28.70 & 55.61 \\

    SRLM -- $\mathcal{M}_3$ \\
    \hdashline
    
    \quad Cat. = 5k & 57.35 & 65.96 & 73.04 & 45.52 & 30.73 & 54.52 \\
    \quad Cat. = 10k & 57.29 & 67.32 & 73.38 & 48.65 & 29.23 & 55.17 \\
    
    \bottomrule
    \end{tabular}
    \end{adjustbox}
    \caption{The results using different size of reasoning catalyst data based on Llama3.1-8B model with 3 epochs optimization on three iterations ($\mathcal{M}_{0,1,2,3}$). We \textbf{bold} the best performance at each iteration.}
    \label{tab:catalyst_effects_same_epochs}
\end{table*}

\begin{table*}[!t]
\setlength{\belowcaptionskip}{0pt}
    \centering
    \begin{adjustbox}{max width=0.8 \textwidth}
    \begin{tabular}{l cccccc}
    \toprule
    \textbf{Data}  & MMLU & GSM8K & ARC-C & HellaSwag & BBH & Avg. \\
    \hline
    IT$_0$ + Cat. & 57.60 & 64.59 & 70.14 & 49.23 & 31.46 & 54.60  \\
    \hline
    
    \multicolumn{7}{l}{\textit{Update $r_j^*$ with $r_j^{t+1}$}} \\
    \hdashline
    
    IT$_1$ + Up. Cat.$_1$ & 59.27 & 66.03 & 74.15 & 49.84 & 31.29 &  56.12 \\
    IT$_2$ + Up. Cat.$_2$ & 57.42 & 65.81 & 71.08 & 47.57 & 29.45 & 54.27 \\
    IT$_3$ + Up. Cat.$_3$ & 55.52 & 65.13 & 69.28 & 49.25 & 30.92 & 54.02 \\
    \hline

    \multicolumn{7}{l}{\textit{Update both $r_j^0, r_j^*$ with $r_j^t, r_j^{t+1}$}} \\
    \hdashline
    IT$_1$ + Up. Cat.$_1$ & 56.99 & 64.75 & 72.87 & 47.94 & 33.83 & 55.28 \\
    IT$_2$ + Up. Cat.$_2$ & 57.18 & 63.99 & 67.92 & 49.78 & 33.11 & 54.40 \\
    IT$_3$ + Up. Cat.$_3$ & 55.09 & 65.13 & 68.17 & 49.42 & 32.57 & 54.08 \\
    
    \bottomrule
    \end{tabular}
    \end{adjustbox}
    \caption{The performance of SRLM with length selector on Llama3.1-8B using \textit{Update} 1k reasoning catalyst data with 1000 steps optimization.}
    \label{tab:update_catalyst}
\end{table*}

\begin{table*}[!t]
\setlength{\belowcaptionskip}{0pt}
    \centering
    \begin{adjustbox}{max width=0.96 \textwidth}
    \begin{tabular}{ll|ccccc|c}
    \toprule
    \textbf{\# Sampling} & \textbf{Methods}  & MMLU & GSM8K & ARC-C & HellaSwag & BBH & Avg. \\
    \hline
    \multirow{2}{*}{1} & Reflection-tuning & 57.08 & 66.49 & 67.32 & 50.34 & 30.09 & 54.26 \\
    & SRLM & 57.91 & 67.25 & 72.70 & 52.88 & 34.06 & 56.96 \\

    \hline
    \multirow{2}{*}{2} & Reflection-tuning & 65.50 & 74.07 & 76.02 & 60.88 & 41.55 & 63.60 \\
    & SRLM & 69.56 & 75.89 & 83.02 & 64.25 & 46.61 & 67.87 \\

    \hline
    \multirow{2}{*}{4} & Reflection-tuning & 71.66 & 80.52 & 82.00 & 70.33 & 52.48 & 71.40 \\
    & SRLM & 78.44 & 83.85 & 89.51 & 76.11 & 57.74 & 77.13 \\

    \hline
    \multirow{2}{*}{8} & Reflection-tuning & 76.42 & 84.38 & 84.56 & 77.81 & 61.09 & 76.85 \\
    & SRLM & 85.41 & 88.32 & 93.94 & 85.46 & 68.23 & 84.27 \\

    \hline
    \multirow{2}{*}{16} & Reflection-tuning & 79.79 & 88.10 & 85.75 & 82.61 & 67.56 & 80.76 \\
    & SRLM & 90.31 & 91.89 & 96.33 & 91.67 & 74.87 & 89.01 \\

    \hline
    \multirow{2}{*}{32} & Reflection-tuning & 82.48 & 90.75 & 88.05 & 86.41 & 72.79 & 84.10 \\
    & SRLM & 93.85 & 93.63 & 98.12 & 95.41 & 80.23 & 92.25 \\

    \hline
    \multirow{2}{*}{64} & Reflection-tuning & 84.80 & 92.27 & 89.16 & 89.24 & 76.27 & 86.35 \\
    & SRLM & 96.15 & 95.38 & 98.72 & 97.62 & 83.31 & 94.24 \\
    
    \bottomrule
    \end{tabular}
    \end{adjustbox}
    \caption{The detailed experimental results with multiple sampling for baseline and SRLM.}
    \label{tab:sampling_times}
\end{table*}

\begin{figure*}
    \centering
    \includegraphics[trim={6cm 5cm 0cm 4cm}, clip, width=1.0\textwidth]{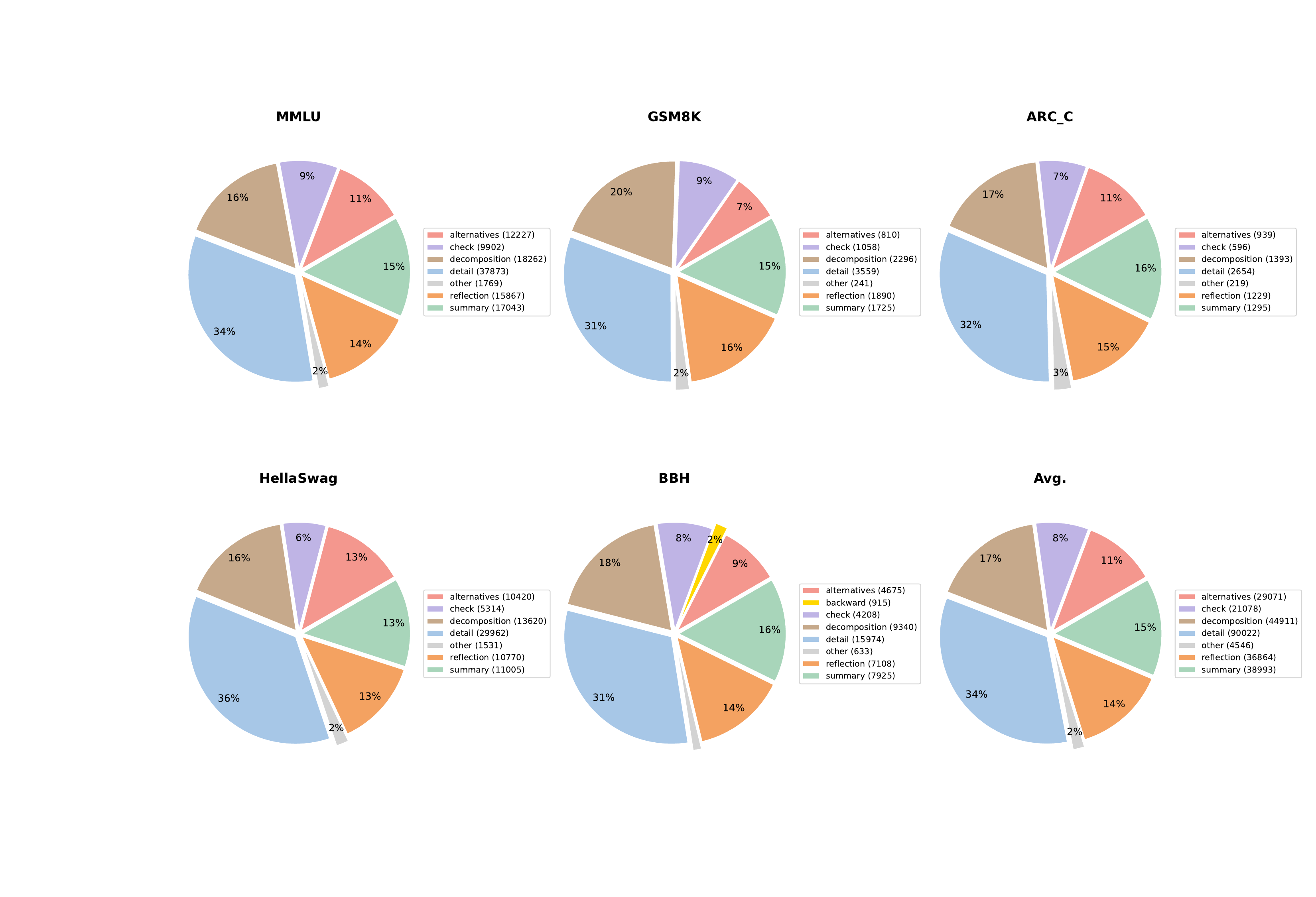}
    \caption{The distribution of meta-reasoning skills on five various benchmarks and the global distribution during inference.}
    \label{fig:all_tags_inference}
\end{figure*}

\subsection{Case Study}

Table~\ref{table:case_study} shows the non-cherry pick test examples of MATH. It is obvious that our proposed SRLM can take advantages of different meta-reasoning skills such as \textit{decomposition}, \textit{detail}, \textit{check}, \textit{reflection} and \textit{summary} to get the correct answer.

\begin{table*}[ht]
\small
    \centering
    \colorbox{babyblue}{
    \begin{tabular}{@{}p{16.2cm}}
    \textbf{Question}: Suppose $a$ and $b$ are positive integers, neither of which is a multiple of 3. Find the least possible remainder when $a^2 + b^2$ is divided by 3. \\\\

    \textbf{Reflection-tuning Answer}: To solve this problem, we need to find the remainder of $a^2 + b^2$ when divided by 3, given that $a$ and $b$ are integers not divisible by 3. \\ First, consider the properties of numbers modulo 3. Any integer $n$ can be expressed as $n = 3k + r$ where $r$ is the remainder when $n$ is divided by 3. Since $a$ and $b$ are not multiples of 3, $r eq 0$. \\ Now, let's express $a$ and $b$ in terms of their remainders when divided by 3: $a = 3m + r_1$ and $b = 3n + r_2$, where $r_1, r_2 eq 0$. \\ Substitute these into $a^2 + b^2$: $ a^2 + b^2 = (3m + r_1)^2 + (3n + r_2)^2 = 9m^2 + 6mr_1 + r_1^2 + 9n^2 + 6nr_2 + r_2^2. $ \\ Notice that all terms containing $m$ and $n$ are multiples of 3, and thus their sum is also a multiple of 3. The only terms not divisible by 3 are $r_1^2$ and $r_2^2$. \\ Since $r_1$ and $r_2$ are not zero, $r_1^2$ and $r_2^2$ can each be 1 or 2 (since they are squares of non-zero remainders when divided by 3). The smallest possible value for $r_1^2 + r_2^2$ is 1 (when both $r_1$ and $r_2$ are 1 or both are 2) and the largest possible value is 4 (when one is 1 and the other is 2). Thus, the remainder of $a^2 + b^2$ when divided by 3 can only be 1 or 2. Therefore, the possible remainders when $a^2 + b^2$ is divided by 3 are \boxed{1, 2}. \\\\

    \textbf{SRLM}: <thoughts> \\ <decomposition> Let's break down the problem into smaller parts to ensure clarity and accuracy in our reasoning. First, we need to understand the behavior of squares modulo 3. Next, we'll examine the possible values of a and b under the constraint that they're not multiples of 3. Finally, we'll compute ( $a^2 + b^2 \mod 3$ ) for these values and determine the remainder. </decomposition> \\ <detail> Understanding the behavior of numbers under modulo 3 involves recognizing that any integer n can be expressed as n = 3q + r, where q is the quotient and r is the remainder. For $n^2$, we have $n^2 = (3q + r)^2 = 9q^2 + 6qr + r^2$. Simplifying gives us $ n^2 \equiv r^2 \pmod{3} $. Thus, the possible remainders for $n^2$ are 0, 1, or 2 depending on $r$. </detail> \\ <detail> Since neither $ a $ nor $ b $ is a multiple of 3, they can take values of 1 or 2 modulo 3. This means $ a \equiv 1 or 2 \pmod{3} $ and similarly for $ b $. We need to evaluate $ a^2 + b^2 $ for these possibilities. </detail> \\ <check> Let's verify each case:- Case 1: $ a \equiv 1 \pmod{3}, b \equiv 1 \pmod{3} $  - Then $ a^2 \equiv 1^2 \equiv 1 \pmod{3} $  - $ b^2 \equiv 1^2 \equiv 1 \pmod{3} $  - So, $ a^2 + b^2 \equiv 1 + 1 \equiv 2 \pmod{3} $ \\ - Case 2: $ a \equiv 1 \pmod{3}, b \equiv 2 \pmod{3} $  - Then $ a^2 \equiv 1^2 \equiv 1 \pmod{3} $  - $ b^2 \equiv 2^2 \equiv 4 \equiv 1 \pmod{3} $  - So, $ a^2 + b^2 \equiv 1 + 1 \equiv 2 \pmod{3} $ \\ - Case 3: $ a \equiv 2 \pmod{3}, b \equiv 1 \pmod{3} $  - Then $ a^2 \equiv 2^2 \equiv 4 \equiv 1 \pmod{3} $  - $ b^2 \equiv 1^2 \equiv 1 \pmod{3} $  - So, $ a^2 + b^2 \equiv 1 + 1 \equiv 2 \pmod{3} $ \\ - Case 4: $ a \equiv 2 \pmod{3}, b \equiv 2 \pmod{3} $  - Then $ a^2 \equiv 2^2 \equiv 4 \equiv 1 \pmod{3} $  - $ b^2 \equiv 2^2 \equiv 4 \equiv 1 \pmod{3} $  - So, $ a^2 + b^2 \equiv 1 + 1 \equiv 2 \pmod{3} $</check>\\<summary> In all cases, the remainder when $ a^2 + b^2 $ is divided by 3 is consistently 2. There were no overlooked scenarios or computational errors, confirming the correctness of the result. </summary>\\ <reflection> By systematically evaluating each case and cross-verifying calculations, I ensured that the solution aligns with the conditions provided. The consistency in outcomes confirms the robustness of the method applied. </reflection> \\ Therefore, the least possible remainder when $ a^2 + b^2 $ is divided by 3 is $\boxed{2}$.</thoughts> \\ 
    \textbf{Based on the above processing, the key reasoning steps can be summarized clearly, followed by the concise answer.} \\ To solve this problem, we need to understand the behavior of squares modulo 3. Any integer $n$ can be expressed as $n = 3q + r$, where $q$ is the quotient and $r$ is the remainder when $n$ is divided by 3. Squaring both sides gives $n^2 = (3q + r)^2 = 9q^2 + 6qr + r^2$. This simplifies to $n^2 \equiv r^2 \pmod{3}$. Therefore, the possible remainders for $n^2$ are 0, 1, or 2 depending on $r$. \\ Since neither $a$ nor $b$ is a multiple of 3, they can be either 1 or 2 modulo 3. We need to calculate $a^2 + b^2$ for these cases: \\ 1. If $a \equiv 1 \pmod{3}$ and $b \equiv 1 \pmod{3}$, then $a^2 \equiv 1^2 \equiv 1 \pmod{3}$ and $b^2 \equiv 1^2 \equiv 1 \pmod{3}$. So, $a^2 + b^2 \equiv 1 + 1 \equiv 2 \pmod{3}$.2. If $a \equiv 1 \pmod{3}$ and $b \equiv 2 \pmod{3}$, then $a^2 \equiv 1^2 \equiv 1 \pmod{3}$ and $b^2 \equiv 2^2 \equiv 1 \pmod{3}$. So, $a^2 + b^2 \equiv 1 + 1 \equiv 2 \pmod{3}$.3. If $a \equiv 2 \pmod{3}$ and $b \equiv 1 \pmod{3}$, then $a^2 \equiv 2^2 \equiv 1 \pmod{3}$ and $b^2 \equiv 1^2 \equiv 1 \pmod{3}$. So, $a^2 + b^2 \equiv 1 + 1 \equiv 2 \pmod{3}$.4. If $a \equiv 2 \pmod{3}$ and $b \equiv 2 \pmod{3}$, then $a^2 \equiv 2^2 \equiv 1 \pmod{3}$ and $b^2 \equiv 2^2 \equiv 1 \pmod{3}$. So, $a^2 + b^2 \equiv 1 + 1 \equiv 2 \pmod{3}$. In all cases, the remainder when $a^2 + b^2$ is divided by 3 is 2. Therefore, the least possible remainder is \boxed{2}. 
    
    \end{tabular}}
    \caption{The (non-cherry-pick) test instance from MATH.}
    \label{table:case_study}
\end{table*}

\end{document}